\documentclass[10pt,twocolumn,letterpaper]{article}

\PassOptionsToPackage{table}{xcolor}

\usepackage[pagenumbers]{cvpr} 

\definecolor{cvprblue}{rgb}{0.21,0.49,0.74}
\usepackage[pagebackref,breaklinks,colorlinks,allcolors=cvprblue]{hyperref}
\usepackage{fontawesome5} 
\usepackage{dashbox}
\def\confName{CVPR}
\def\confYear{2027}

\title{\datasetstyle: A Benchmark for Physical Reasoning \\in Vision-Language-Action Models via Dynamic Extrinsic Dexterity}

\author{
Hao Vo\textsuperscript{1*},
Taisei Hanyu\textsuperscript{1*},
Yuki Ikebe\textsuperscript{1*},
Pham Tri Quang\textsuperscript{2}, \\
Sreevenkata Anjani Tishita Godavarthi\textsuperscript{1},
Nghi D. Q. Bui\textsuperscript{3},
Daniel Sonntag\textsuperscript{4}, \\
Chase Rainwater\textsuperscript{1},
Duy Minh Ho Nguyen\textsuperscript{4,5,6},
Anh Nguyen\textsuperscript{7},
Ngan Le\textsuperscript{1} \\
\textsuperscript{1}AICV Lab, University of Arkansas, USA \quad
\textsuperscript{2}VinUniversity, Vietnam \\
\textsuperscript{3}Center for AI Research, VinUniversity, Vietnam \quad
\textsuperscript{4}University of Stuttgart, Germany \\
\textsuperscript{5}DFKI, Germany \quad
\textsuperscript{6}Max Planck Research School for Intelligent Systems, Germany \\
\textsuperscript{7}University of Liverpool, UK \\[2pt]
\textsuperscript{*}Equal contribution \\[2pt]
{%
\ttfamily\small
\makebox[\linewidth][c]{%
\parbox{0.95\linewidth}{\centering
\href{https://uark-aicv.github.io/PhyDEX/}{\faGlobe\ Project Page}
}}%
}%
}

\usepackage[toc,page,header]{appendix}
\usepackage{minitoc}
\usepackage{epigraph}
\usepackage{caption} 
\definecolor{mylime}{RGB}{5,129,69}
\usepackage{soul}
\setulcolor{mylime}
\setul{}{0.68pt}
\usepackage{pifont}
\definecolor{darkgreen}{RGB}{0,128,0} 
\definecolor{lightblue}{rgb}{0.9,0.95,1.0} 
\newcommand{\dataset}{\textit{PhyDEX}\xspace}

\usepackage{booktabs}  
\usepackage{multirow}  
\usepackage{graphicx}  
\usepackage{makecell}  
\definecolor{dexorange}{RGB}{255,179,102}

\newcommand{\datasetstyle}{\textit{\textbf{Phy\textcolor{dexorange}{DEX}}}\xspace}

\usepackage{caption}   
\makeatletter
\g@addto@macro\@maketitle{%
  \vspace{-0.5em}%
  \begin{center}
    \captionsetup{type=figure}
    \includegraphics[width=\textwidth]{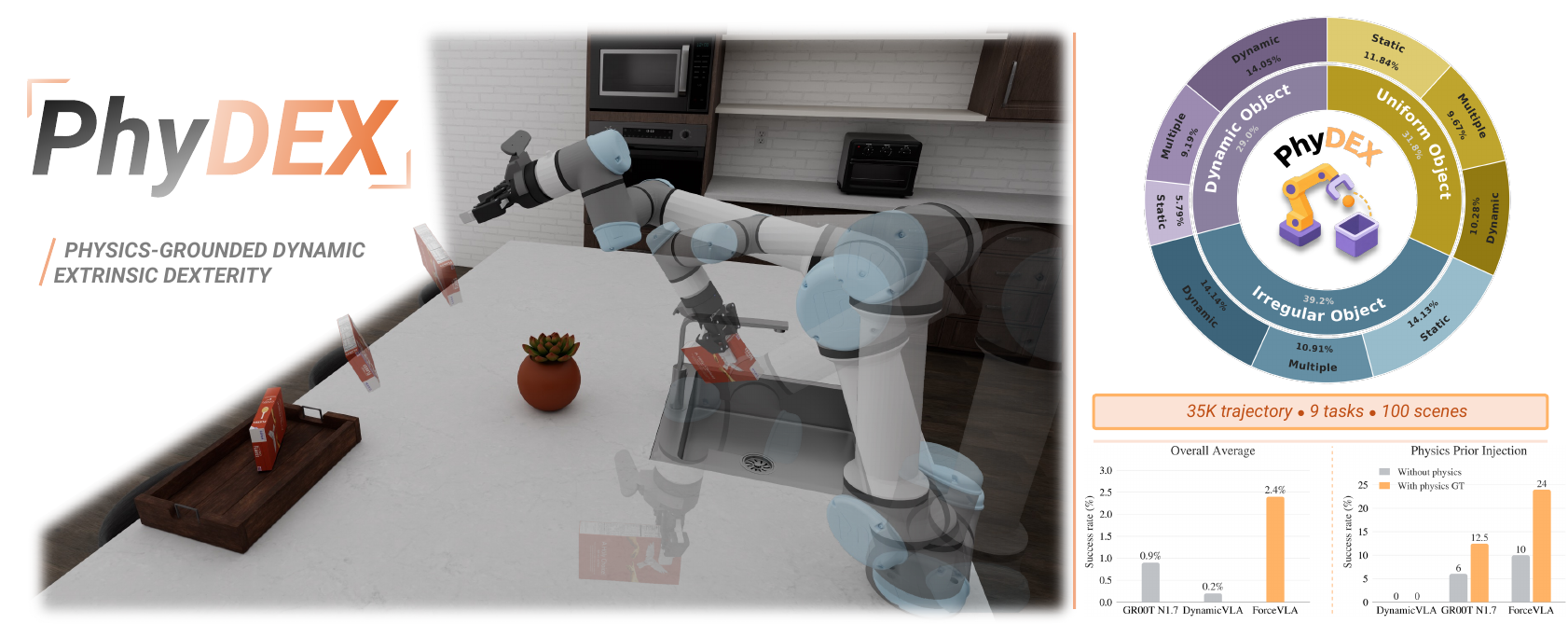}
    \caption{\textbf{Overview of \datasetstyle.} \textit{Left:} a robot arm throws an object into a target destination, which requires reasoning about the object's mass, center of mass, and friction. \textit{Right:} \datasetstyle contains 35K trajectories across 9 tasks and 100 scenes, covering uniform, irregular, and dynamic objects (top). Current VLA policies all reach overall success rates below 3\% (bottom left), while injecting ground-truth physical properties substantially improves performance when throwing uniform objects to a static single destination (bottom right).}
    \label{fig:teaser}
  \end{center}%
  \vspace{0.5em}%
}
\makeatother

\begin{document}
\maketitle
\doparttoc              
\faketableofcontents

\begin{abstract}
Vision-Language-Action (VLA) models have advanced rapidly in language-conditioned robotic manipulation, yet existing benchmarks mainly evaluate gripper--object contact and largely ignore how objects interact with the environment through gravity, friction, inertia, and mass distribution. As a result, it remains unclear whether current policies truly understand physics or merely interpolate from their training data. We address this gap through the lens of dynamic extrinsic dexterity and introduce \datasetstyle, a throwing benchmark in which a robot must infer an object's mass, center of mass, and friction to send it to a destination beyond its kinematic workspace. \datasetstyle contains 35K trajectories (143 hours) across nine tasks and 100 scenes, covering uniform, irregular, and dynamic-mass objects with static, multiple, and moving destinations. All data are generated by a scalable, fully automatic pipeline that combines grasp generation, motion planning, and closed-loop throw optimization under randomized physical properties. We evaluate three representative VLAs, GR00T N1.7, DynamicVLA, and ForceVLA, and find that all achieve success rates below 3\%. Our analysis reveals two key bottlenecks: limited robustness of pretrained models, which causes most failures before the throw, and physical blindness, as providing ground-truth physical properties roughly doubles success. Experiments on a humanoid robot and in the real world show the same failure patterns. We hope \datasetstyle will drive the development of physically grounded robot policies.
\end{abstract}    
\section{Introduction}
\label{sec:intro}



Physical understanding has widely recognized as a cornerstone of language-conditioned robotic manipulation systems~\cite{bousetouane2025physical, ye2026world, han2026geometric, pi2026pi07}. A broad spectrum of methods have emerged, ranging from hybrid frameworks~\cite{xiao2026enpire,liang2023code} to unified systems~\cite{pi2026pi07,kim2024openvla,zitkovich2023rt}, from vision-language architectures~\cite{zhang2026dreamvla, bjorck2025gr00t,kim2026cosmos} to multi-sensory models~\cite{yu2026forcevla,niu2026t}, and from static methods~\cite{zhang2026dreamvla, bjorck2025gr00t,kim2026cosmos,qian2026geopredict,han2026geometric,qu2025spatialvla,li2026bridgevla,li2025spatial} to dynamic paradigm~\cite{zhang2025dynamic,li2025sim, liu2026rdt2,huang2026rt,tang2025vlash}.

\begin{figure*}[t]
    \centering
    \includegraphics[width=\linewidth]{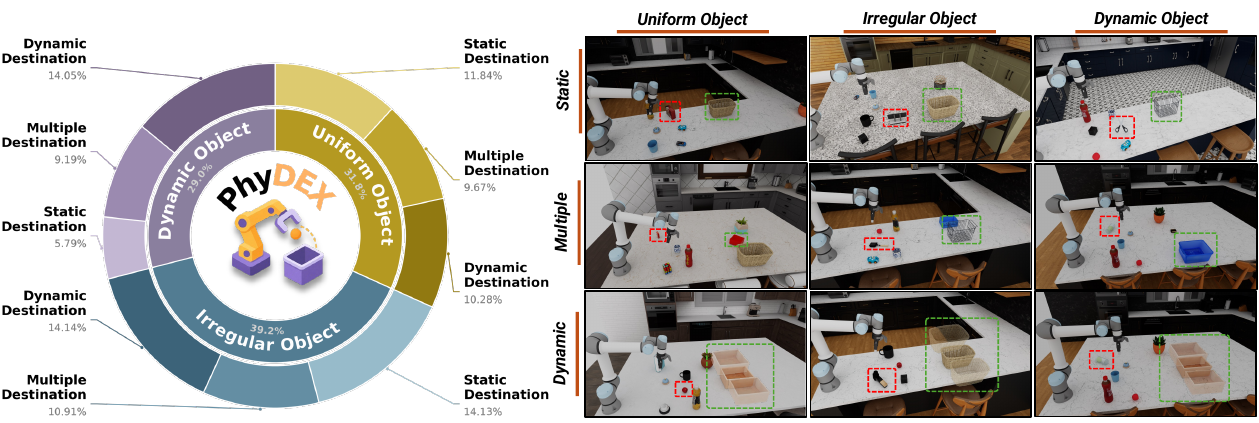}
\caption{\textbf{Task taxonomy and statistics of \datasetstyle.} 
\textit{Left:} distribution of episodes across three object types (inner ring) and three destination types (outer ring). 
\textit{Right:} example scenes for each of the nine object--destination combinations. \colorbox{red!15}{boxes} mark the target object and \colorbox{green!15}{boxes} mark the destination. }

\label{fig:statistic}
    \vspace{-4mm}
\end{figure*}

%
Despite this great progress, recent works predominantly focus on the intrinsic gripper–object contact, whether in standard manipulation tasks such as \textit{Pick-and-Place}, \textit{Re-orientation}, and \textit{Opening \& Closing Articulated Objects}~\cite{liu2023libero,nasiriany2024robocasa,chen2025robotwin,zhang2025vlabench}, or in contact-rich tasks such as \textit{Insertion}, \textit{Fragile Grasping}, and \textit{Wiping}~\cite{fang2026force,yu2026forcevla,huang2025tactile,li2026forcevla2,kim2026rldx,liu2024forcemimic,fang2023rh20t,niu2026t,shan2026learning}. However, they largely overlook how the manipulators interacts with its surrounding environment (e.g., gravity, friction, inertia, aerodynamics, mass distribution, etc.), namely the newtonian mechanics underlying everyday physical interactions, which substantially limits their ability to reflect genuine physical intelligence.\\
To fill this gap, we recast language-conditioned robotic manipulation through the lens of dynamic extrinsic dexterity~\cite{dafle2014extrinsic}, focusing on \textit{throwing-based object manipulation}. In this setting, the manipulator must estimate the object's physical properties, such as its mass, center of mass (CoM), and friction, and plan the forces and contacts needed to send the object to a target location outside its own kinematic workspace.\\
We introduce \datasetstyle, a high-fidelity synthetic manipulation dataset comprising 35k trajectories across 9 throwing tasks and 100 scenes. It covers rigid, irregular, and dynamic-mass objects, as well as diverse destination types, with all physical properties randomized. \dataset is generated by a fully decoupled, autonomous simulation pipeline that separates asset specification, skill policies, task composition, and rendering. Each task is built by retrieving embodiments, scenes, and objects from an asset library and composing scripted skill policies into complete behaviors. This compositional design enables automatic generation of object-level annotations, physical randomization, and collision-aware motion plans, supporting systematic VLA validation through fine-tuning and closed-loop evaluation.

To evaluate the physical understanding of current policies, we select three representative policies: GR00T N1.7~\cite{gr00t-n1} as a general-purpose policy, DynamicVLA~\cite{xie2026dynamicvla} as a high-frequency policy, and ForceVLA~\cite{yu2026forcevla} as a multi-sensory VLA. Our experiments show that current policies are not yet reliable for tasks requiring physical understanding, with average success rates below 10\%. We further identify generalization and the probing of object physical properties as the main bottlenecks of current systems. To confirm that this trend holds beyond a single setup, we additionally conduct evaluations across multiple platforms, including a humanoid robot and real-world experiments.

In summary, our contributions are threefold:

\begin{itemize}

\item We propose \datasetstyle, a diverse and comprehensive dataset of nine throwing tasks for studying physical understanding in robot manipulation.
\item We introduce a scalable construction pipeline that generates diverse, end-to-end data for throwing tasks.
\item We conduct extensive evaluations across multiple embodiments, including robot arms and humanoids, in both simulation and real-world experiments. Our results show that data scarcity and physical blindness are the main bottlenecks keeping current policies from performing reliably on physical understanding tasks.

\end{itemize}

\section{Related Works}

\subsection{Language-conditioned Robot Manipulation} Recently, language-conditioned robot manipulation systems have attracted much attention and achieved great progress, owing to their generalizability, scalability, and potential for deployment in the physical world. Related methods can be roughly categorized into two branches: autoregressive-based approaches~\cite{brohan2022rt,kim2024openvla,li2025unified,li2024manipllm,pertsch2025fast,wen2025tinyvla,goyal2025vla} and diffusion-based approaches~\cite{bjorck2025gr00t,pi2026pi07,li2024cogact,wen2025diffusionvla,ye2026world,kim2026cosmos,han2026geometric}. While these methods have significantly advanced overall performance, their physical knowledge is interpolated from the training data distribution and its reliability remains unverified. 

\subsection{Language-conditioned Dynamic Manipulation}
Our work is closely related to dynamic manipulation, which demands policies that are fast, reactive, and precise. Modeling such dynamics is challenging, as it relies on physical properties of both the objects and the manipulator that are difficult to estimate~\cite{mason1993dynamic}. In the language-conditioned setting, most recent work tracks a target object under static and predictable conditions~\cite{zhang2025dynamic,li2025sim}, adopts asynchronous inference to hide latency~\cite{liu2026rdt2,huang2026rt,tang2025vlash}, or extends control to more precise 6-DoF manipulation~\cite{liao2026dynamicmanip,xie2026dynamicvla}. While effective, these methods are evaluated primarily on kinematic accuracy. They optimize for trajectory matching and leave the understanding of physical properties implicit, so a failure cannot be attributed to kinematic error or to a misestimated physical property. \textit{In this work, we design a beyond-reach manipulation benchmark in which the object must be moved to a target location outside the kinematic range of the manipulator, making the understanding of physical properties a decisive factor for success.}

\subsection{Physical Property Understanding} To throw an object accurately to a target, a robot must first estimate the object's physical properties. Studies in cognitive science~\cite{yuan2017connecting,mcadams1993recognition} suggest that humans infer such properties not only from visual appearance but also from material experience. Existing approaches to physical property estimation fall into two main lines of work: those that encode an object's properties implicitly~\cite{wu2016physics,bauza2019omnipush,purri2020teaching,yuan2017connecting} and those that predict them explicitly~\cite{liu2026vitacphys,yu2024octopi,sundaram2019learning,standley2017image2mass}. While effective, these methods mostly target object-centric settings in general domains, and robotic scenarios remain underexplored. A few works in robot learning~\cite{liu2026vitacphys,xie2024deligrasp} have tackled this problem, but they are limited to simple pick-and-place or contact-rich manipulation tasks. These tasks are not demanding enough to properly evaluate a model's physical understanding. \textit{To fill this gap, we explicitly randomize object physical properties, such as mass, CoM, and friction, to evaluate a model's ability to accurately estimate object properties and reason about physics.}

\begin{figure*}[t]
    \centering
    \includegraphics[width=\linewidth]{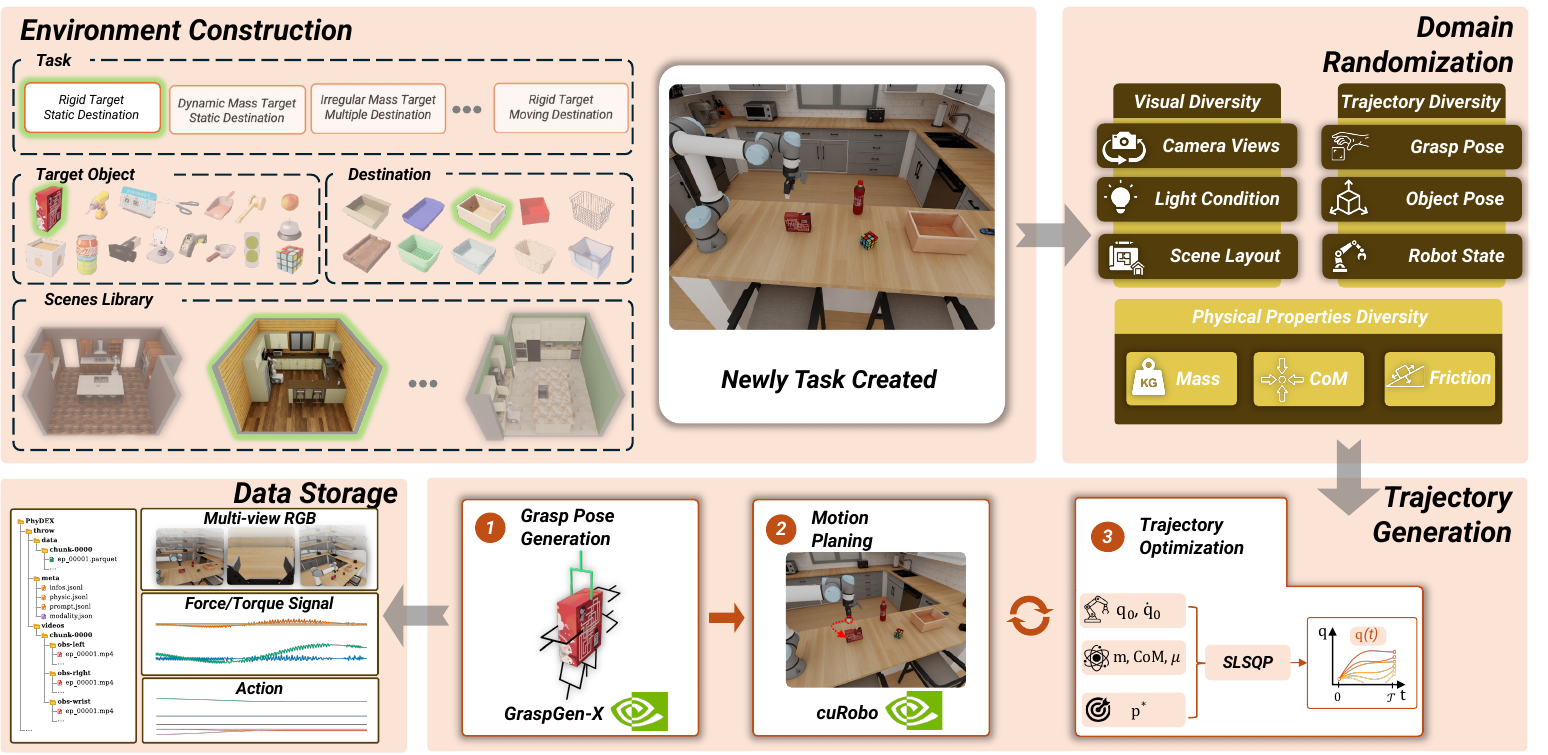}
\caption{\textbf{Overview of the \datasetstyle data generation pipeline.} 
\textbf{(a) Environment construction:} given a task template, we retrieve a target object, a destination, and a scene from the asset library to create a new task. 
\textbf{(b) Domain randomization:} we diversify each task visually (camera views, lighting, scene layout), in its trajectories (grasp pose, object pose, robot state), and in its physical properties (mass, CoM, friction). 
\textbf{(c) Trajectory generation:} GraspGen-X proposes grasp poses, cuRobo plans a collision-free approach, and an SLSQP-based optimizer computes the throwing trajectory $q(t)$ from the initial joint state $(q_0, \dot{q}_0)$, the object properties $(m, \text{CoM}, \mu)$, and the target $p^*$, refining it in a closed loop until the throw succeeds. 
\textbf{(d) Data storage:} each successful episode is saved in LeRobot format with multi-view RGB, force/torque signals, and actions.}
\label{fig:pipeline}
    \vspace{-4mm}
\end{figure*}

\section{\datasetstyle}

In this section, we present the task taxonomy, detailed dataset statistics, and the data creation pipeline, which are illustrated in Figure \ref{fig:statistic} and Figure \ref{fig:pipeline}, respectively.

\subsection{Task Taxonomy \& Statistics}

\dataset{} comprises 9 tasks, 35{,}357 episodes, and 143.2 hours of data recorded at 10\,Hz. It is designed to evaluate how well policies estimate the physical properties of objects from different perspectives.

\noindent\textbf{Tasks.} In \dataset{}, policies perform a throwing task at three levels of object difficulty:
(1) \textit{Uniform Object} (31.8\%): objects with uniformly distributed mass and a centered center of mass (CoM);
(2) \textit{Irregular Object} (39.2\%): objects with non-uniform mass distribution and a CoM offset toward one end; and
(3) \textit{Dynamic Object} (29.0\%): objects whose mass and CoM change during the interaction.
In addition, to test generalization and timing, we use three destination settings: \textit{static single destination} (31.76\%), \textit{static multiple destinations} (29.77\%), and \textit{moving single destination} (38.47\%).

\noindent \textbf{Assets.} We provide 107 curated object assets, sourced from RoboCasa~\cite{nasiriany2024robocasa}, RoboTwin 2.0~\cite{chen2025robotwin}, and Objaverse-XL~\cite{deitke2023objaverse}: 20 destination objects, 46 uniform objects, 37 irregular objects, and 4 dynamic objects. For environments, \dataset{} includes 100 RoboCasa kitchen scenes spanning diverse layouts and styles.

\subsection{Benchmark Construction Pipeline}
In this section, we describe the data synthesis pipeline of \dataset{}.
We first construct the environment by specifying the task, target object,
destination, and scene. We then apply domain randomization to diversify
these environments. Finally, we generate approaching and throwing
trajectories and define the procedures for storing the resulting data.

\subsubsection{Environment Construction} 
As illustrated in Figure~\ref{fig:pipeline}, given a task description template, we first retrieve the task-relevant assets from the asset library: the target object, the destination object, and the scene. We use a UR5e arm equipped with a Robotiq 2F-85 gripper as the robot embodiment, represented as a stable USD asset. The scene assets are room-scale environments with diverse layouts and styles, each annotated with fine-grained manipulation areas to support natural, context-aware task construction. Our object library spans three mass categories (uniform, irregular, and dynamic) and covers a range of physical properties, sizes, and layouts. Objects are sampled equally across categories to keep the benchmark balanced. We also place distractor objects around the target object to encourage policy generalization. Destination objects likewise vary in motion direction, size, and layout. All functional annotations are managed through a unified interface, which makes asset configuration straightforward while keeping physical fidelity high.

\subsubsection{Domain Randomization} 
To enrich the diversity of \dataset{}, we apply three types of randomization.

\noindent \textbf{Visual diversity.} We perturb the poses of the two side cameras and the wrist camera by up to $\pm5^\circ$ in rotation and $\pm5$~cm in translation. We also build a library of 100 environment maps, each with randomized light temperature and intensity, to simulate diverse natural illumination conditions.

\noindent \textbf{Trajectory diversity.} We introduce additional randomness in grasp pose selection output from GraspGen-X~\cite{han2026graspgen}. For instance, our autonomous grasp pose generation pipeline produces millions of grasp candidates. We retain the top 64 high-confidence grasps that are geometrically diverse, requiring each selected grasp to differ from all previously selected ones by more than 1~cm in position or $20^\circ$ in orientation, and randomly sample the final grasp pose from this set. We further randomize the object position by $\pm10$~cm along the $x$- and $y$-axes, the object yaw by $\pm180^\circ$, and the initial robot joint positions by $\pm0.09$~rad.

\noindent \textbf{Physical diversity.} To evaluate whether policies can infer physical properties, we randomize each object's physical parameters independently of its appearance, so visually identical objects may behave differently. Specifically, we scale the object mass by a factor in $[0.5, 2.0]$, offset the center of mass by up to $\pm15$~mm along each axis, and scale the friction coefficient by a factor in $[0.8, 1.2]$.

\subsubsection{Trajectory Generation} 
To produce plausible and successful trajectories, we introduce a closed-loop optimization procedure that executes each candidate trajectory end-to-end and iteratively refines it based on the resulting feedback. The procedure consists of three main steps. 

\noindent \textbf{Grasp generation.} As described above, we pre-compute 64 grasp candidates for each object. Given the object's placement pose in the scene, we first discard grasps that would collide with surrounding geometry, such as the supporting surface or nearby objects. The remaining feasible grasps are then ranked by confidence score and geometric diversity, and one is sampled from the top-ranked candidates and passed to the motion planning stage.  

\noindent \textbf{Motion planning.} We then provide cuRobo~\cite{sundaralingam2023curobo} with the scene point cloud extracted from the simulator and the target grasp pose, and it generates a collision-free trajectory that approaches the object, grasps it and lifts it. 

\begin{table}[t]
\centering
\caption{\textbf{Performance comparison of three baseline VLA models on \datasetstyle} under full diversity ($\alpha = 1$).}
\label{tab:main_results}
\renewcommand{\arraystretch}{1.25}
\resizebox{\columnwidth}{!}{
\begin{tabular}{c|lccc}
\toprule
 & Destination & GR00T N1.7~\cite{bjorck2025gr00t} & DynamicVLA~\cite{xie2026dynamicvla} & ForceVLA~\cite{yu2026forcevla} \\
\midrule

\multirow{3}{*}{\rotatebox[origin=c]{90}{\makecell{Uniform}}}
 & Single   & 6\% & 0\% &  10\% \\
 & Multiple & 2\% & 0\% & 10\% \\
 & Dynamic  & 0\%  & 0\%  & 0\%   \\
\midrule

\multirow{3}{*}{\rotatebox[origin=c]{90}{\makecell{Irregular}}}
 & Single   & 0\% & 0\% & 0\% \\
 & Multiple & 0\% & 2\% & 0\% \\
 & Dynamic  & 0\%  & 0\%  & 0\%   \\
\midrule

\multirow{3}{*}{\rotatebox[origin=c]{90}{\makecell{Dynamic}}}
 & Single   & 0\% & 0\% & 0\% \\
 & Multiple & 0\% & 0\% & 2\% \\
 & Dynamic  & 0\%  & 0\%  & 0\%   \\
\midrule

\multicolumn{2}{l}{\textit{Overall Average}} & 0.9\% &  0.2\% & 2.4\% \\
\bottomrule
\end{tabular}
}
\vspace{-4mm}
\end{table}

\noindent \textbf{Throw optimization.} After lifting, a throw planner takes the measured joint state, grasp offset, and the object's mass, CoM, and inertia. It assumes a rigid grasp until release and drag-free ballistic flight afterward. It jointly optimizes the release joint configuration $\mathbf q_r$, the release joint velocities $\dot{\mathbf q}_r$, the throw duration $T$, and the flight time $\tau$. Each joint follows a quintic polynomial set by the measured start state, the release state $(\mathbf q_r, \dot{\mathbf q}_r)$, and zero acceleration at both ends, so these variables define the whole motion. Forward kinematics and the Jacobian give the object's release position and velocity:
\begin{equation}
\mathbf p_r = f(\mathbf q_r), \qquad \mathbf v_r = J(\mathbf q_r)\,\dot{\mathbf q}_r ,
\end{equation}
which must satisfy the ballistic interception constraint
\begin{equation}
\mathbf p_r + \tau \mathbf v_r + \tfrac{1}{2}\mathbf g \tau^2 = \mathbf p_{\mathrm{dst}}(T+\tau) + (\delta_x, \delta_y, 0)^{\mathsf T},
\end{equation}
where $\mathbf p_{\mathrm{dst}}(T+\tau)$ is the destination opening at arrival time, so moving targets are handled, and $\boldsymbol\delta$ is a horizontal aim offset that starts at zero. Starting from an IK-based initial guess, SLSQP minimizes joint velocities, accelerations, and duration. Along the trajectory it enforces joint, velocity, torque, and table-clearance limits, plus a descending arrival into the opening. A plan is accepted only if it satisfies Eq.~(2) within 3\,mm and all constraints at every physics step.

 The plan runs in simulation, with the gripper opening timed to the control clock. Effects the planner does not fully model cause the executed release to differ from the plan: tracking error under the payload's mass and CoM, friction-dependent slip in the grasp, and release delay. After a miss, we predict the horizontal miss $\mathbf e_k$ from the measured release state and update the aim offset as $\boldsymbol\delta_{k+1} = \boldsymbol\delta_k - 0.5\,\mathbf e_k$, then re-plan. If no feasible throw remains, a new grasp is sampled. An episode is kept only when the simulator reports success.

\subsubsection{Data Storage}
Each successful trajectory is stored in the standard LeRobot format~\cite{cadene2026lerobot}. A record contains the language instruction, multi-view RGB images with their camera parameters, the robot's proprioceptive states, and the action labels. It also holds a 2\,s window of wrist force/torque (F/T) readings, which supports training multi-sensory VLA models such as ForceVLA~\cite{yu2026forcevla}. Other modalities, such as depth maps, grounding annotations and bounding boxes, can be added to the recording with a simple configuration option. For each trajectory we also save an HDF5 file holding the scene metadata and a replayable copy of the trajectory, so the data can be reused and extended later.

\section{Experiments}

\hl{}

\begin{figure}[t]
    \centering
    \includegraphics[width=0.8\columnwidth]{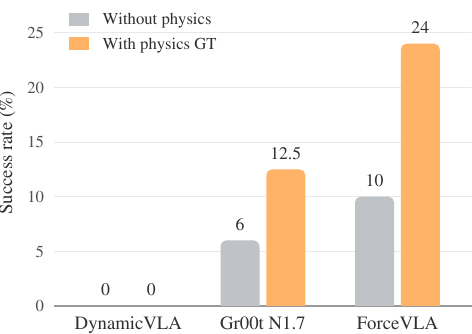}
\caption{\textbf{Effect of ground-truth physical properties.} Success rates with and without access to the object's ground-truth physical properties. }
\label{fig:physics_gt_comparison}
    \vspace{-6mm}
\end{figure}
\subsection{Experimental Setup}
We evaluate three representative policies, each chosen to test a different hypothesis. First, we evaluate GR00T N1.7~\cite{gr00t-n1}, a general-purpose VLA, to examine whether a model that claims broad physical intelligence can actually perform this task. Second, we evaluate DynamicVLA~\cite{xie2026dynamicvla} to test whether a highly reactive policy performs well on this dynamic manipulation problem. Finally, we include ForceVLA~\cite{yu2026forcevla} to assess whether simply adding a force sensor to the policy is enough to give the model physical intelligence. Following standard robot manipulation benchmarks~\cite{chen2025robotwin,nasiriany2024robocasa,liu2023libero}, we report \textit{success rate} ($\mathcal{SR}$) as our evaluation metric. To ensure reproducible and reliable results, we fine-tune a separate model for each task and evaluate each policy over 50 rollouts.

\begin{table}[t]
\centering
\caption{\textbf{Effect of benchmark diversity.} Success rates on static single-destination tasks under different diversity scores $\alpha$.}
\renewcommand{\arraystretch}{1.25}
\resizebox{\columnwidth}{!}{
\begin{tabular}{c|lccc}
\toprule
$\alpha$ & Object & GR00T N1.7~\cite{bjorck2025gr00t} & DynamicVLA~\cite{xie2026dynamicvla} & ForceVLA~\cite{yu2026forcevla} \\
\midrule
\multirow{4}{*}{1}   & Uniform   & 6\% & 0\% & 10\% \\
                     & Irregular & 0\% & 0\% & 0\%  \\
                     & Dynamic   & 0\% & 0\% & 0\%  \\
\cmidrule(lr){2-5}
                     & \textit{Average} & 2\% & 0\% & 3.3\% \\
\midrule
\multirow{4}{*}{0.5} & Uniform   & 6\% & 2\% & 16\% \\
                     & Irregular & 6\% & 4\% & 10\% \\
                     & Dynamic   & 2\% & 0\% & 8\% \\
\cmidrule(lr){2-5}
                     & \textit{Average} & 4.7\% & 2\% & 11.3\% \\
\midrule
\multirow{4}{*}{0}   & Uniform   & 18\% & 6\% & 44\% \\
                     & Irregular & 10\% & 4\% & 40\% \\
                     & Dynamic   & 0\% & 2\% & 22\% \\
\cmidrule(lr){2-5}
                     & \textit{Average} & 9.3\% & 4\% & 35.3\% \\
\bottomrule
\end{tabular}
}

\label{tab:diversity}
    \vspace{-6mm}
\end{table}

\subsection{Implementation Details}

We build our data generation and rendering pipeline on NVIDIA Isaac Sim~\cite{isaacsim}, a GPU-accelerated simulator that provides photorealistic rendering and accurate physics. Data generation runs on 8 NVIDIA RTX A6000 GPUs with 128 parallel environments. The physics simulation runs at 120~Hz, and data are recorded at 10~Hz. Each RGB camera captures images at a resolution of $352 \times 256$. The robot state consists of the 6 joint angles and the gripper open/close state. Wrist force/torque readings are stored over a 2~s window. We fine-tune every policy on 16 NVIDIA A100 (40GB) GPUs using the default hyperparameters released by its authors.

\subsection{Overall Benchmark Results}
\label{sec:overall_results}
As shown in Table~\ref{tab:main_results}, all three policies perform poorly on our benchmark, with overall success rates below 10\%. Most successes occur on objects with uniform mass and static destinations, and no policy completes a single task with a dynamic destination. When objects have more challenging physical properties, such as non-uniform or dynamic mass, the policies rarely complete a full trajectory. The few successes by DynamicVLA and ForceVLA are likely due to chance rather than real competence. We attribute these failures to two main factors. \textit{(1) Limited robustness of pretrained models:} Current policies struggle to generalize to our highly diverse data, and only about 21.24\% of trajectories include a successful grasp of the object. This is most pronounced for DynamicVLA, which builds on the lightweight SmolLM2~\cite{allal2025smollm2} backbone to reach 25\,Hz inference. However, this backbone is too weak for out-of-distribution tasks, and DynamicVLA grasps the object in only 3.57\% of its trials. The other policies achieve higher grasping rates but still struggle under our domain randomization, which we discuss further in the following sections. \textit{(2) Physical blindness in policy design:} Although an average of 21.24\% of trajectories reach the throwing stage, the overall success rate is only about 1.2\%, meaning that just 5.6\% of throws succeed. This result shows that current models have not yet achieved a reliable understanding of physics, since these physical properties cannot be directly observed through images, language instructions, or additional sensors. We therefore believe that future policies require explicit mechanisms for probing and inferring such physical properties.

\begin{table}[t]
\centering
\caption{\textbf{Cross-embodiment evaluation on a humanoid robot.} \textit{Top:} humanoid scenes. \textit{Bottom:} success rates by object type.}
\includegraphics[width=\columnwidth]{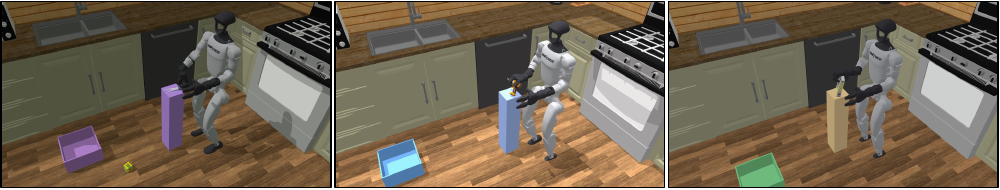}

\vspace{2mm}

\renewcommand{\arraystretch}{1.25}
\resizebox{\columnwidth}{!}{
\begin{tabular}{lccc}
\toprule
Object & GR00T N1.7~\cite{bjorck2025gr00t} & DynamicVLA~\cite{xie2026dynamicvla} & ForceVLA~\cite{yu2026forcevla} \\
\midrule
Uniform   & 10\% & 2\% & 14\% \\
Irregular & 8\%  & 0\% & 6\%  \\
Dynamic   & 0\%  & 0\% & 6\%  \\
\midrule
\textit{Average} & 6\%  & 0.7\% & 8.7\%  \\
\bottomrule
\end{tabular}
}

\label{tab:cross_embod}
    \vspace{-4mm}

\end{table}

\subsection{Further Analysis}

\noindent\textbf{Effect of Benchmark Diversity.} As shown in Table~\ref{tab:main_results}, most policies achieve near-zero success rates on \dataset. We attribute this mainly to the high diversity of our benchmark. Existing benchmarks either apply limited domain randomization, as in LIBERO~\cite{liu2023libero}, or evaluate on a small subset of their data. For example, RoboSPA~\cite{fan2026robospa} provides 527K trajectories but evaluates on only 250 trajectories per task. Such settings can produce non-zero success rates that hide how poorly pretrained policies generalize. To test this explanation, we introduce a diversity score $\alpha \in [0, 1]$ that controls the strength of domain randomization. Each randomized parameter is normally sampled from a range $[a, b]$, and we scale the width of this range by $\alpha$ around its nominal value. Thus $\alpha = 1$ recovers the full \dataset setting and $\alpha = 0$ disables randomization entirely. As shown in Table~\ref{tab:diversity}, success rates rise steadily as diversity decreases. ForceVLA improves the most, from 3.3\% at $\alpha = 1$ to 11.3\% at $\alpha = 0.5$ and 35.3\% at $\alpha = 0$, a more than tenfold gain. GR00T N1.7 follows the same trend, rising from 2\% to 4.7\% and then 9.3\%, while DynamicVLA stays below 4\% in every setting, which again points to the limited capacity of its lightweight backbone. These results confirm that the high diversity of \dataset is a major reason for the near-zero scores, and that current pretrained policies are far less robust than their results on less diverse benchmarks suggest. However, even without any randomization, no policy exceeds 44\% on any object type, and performance remains lowest on dynamic objects, where GR00T N1.7 still fails completely. This shows that reducing diversity alone does not solve the task, and that inferring physical properties remains a core challenge.

\noindent\textbf{Effect of Physical Properties.} As discussed above, we hypothesize that current policies need a better way to infer the physical properties of objects in order to complete these tasks. To test this hypothesis, we extract the 
ground-truth physical properties of each object and give them to the policy by concatenating them with the robot proprioception embedding. As shown in Figure~\ref{fig:physics_gt_comparison}, adding this physical prior clearly improves performance, raising $\mathcal{SR}$ by 12.5\% for GR00T N1.7 and 24\% for ForceVLA. DynamicVLA's performance stays unchanged, which we attribute to its limited pretraining. These results support our hypothesis. We believe this finding gives the community evidence to motivate new designs, such as combining VLAs with reinforcement learning, injecting VLM priors, or adding richer sensor information.

\noindent\textbf{Cross-Embodiment Evaluation.} To check whether these failures are specific to our UR5e arm, we build a small cross-embodiment dataset on a Unitree humanoid using the same generation pipeline. It contains 2,400 episodes with static destinations: 640 for uniform objects, 1,280 for irregular objects, and 480 for dynamic objects. We evaluate each policy over 50 rollouts. As shown in Table~\ref{tab:cross_embod}, DynamicVLA fails on nearly every task. Among the vision-only policies, GR00T N1.7 performs best, but its average success rate is only 6\%. ForceVLA, which also uses force/torque input, reaches 8.7\%, and thanks to the simpler scene setup it also succeeds on some dynamic-object trials (6\%). Overall, the failure patterns on the humanoid closely match those on \dataset, which suggests these failures are shared across embodiments and are not caused by a single robot platform.

\begin{figure}[t]
    \centering
    \includegraphics[width=\columnwidth]{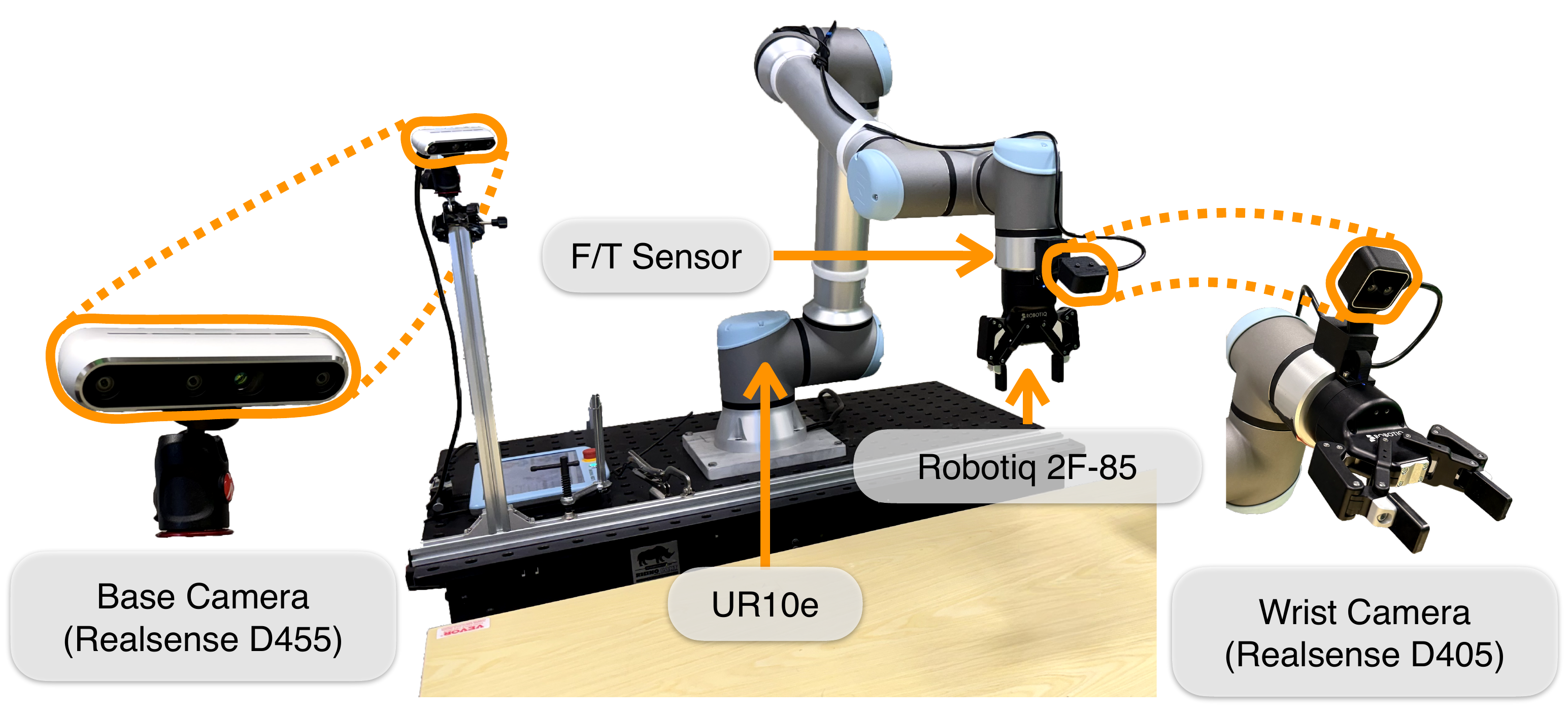}
\caption{\textbf{Real-world experimental setup.} A UR10e arm with a Robotiq 2F-85 gripper and a wrist-mounted F/T sensor, observed by a base camera (RealSense D455) and a wrist camera (RealSense D405).}
\label{fig:robot_setup}
\end{figure}

\begin{figure}[t]
    \centering
    \includegraphics[width=\columnwidth]{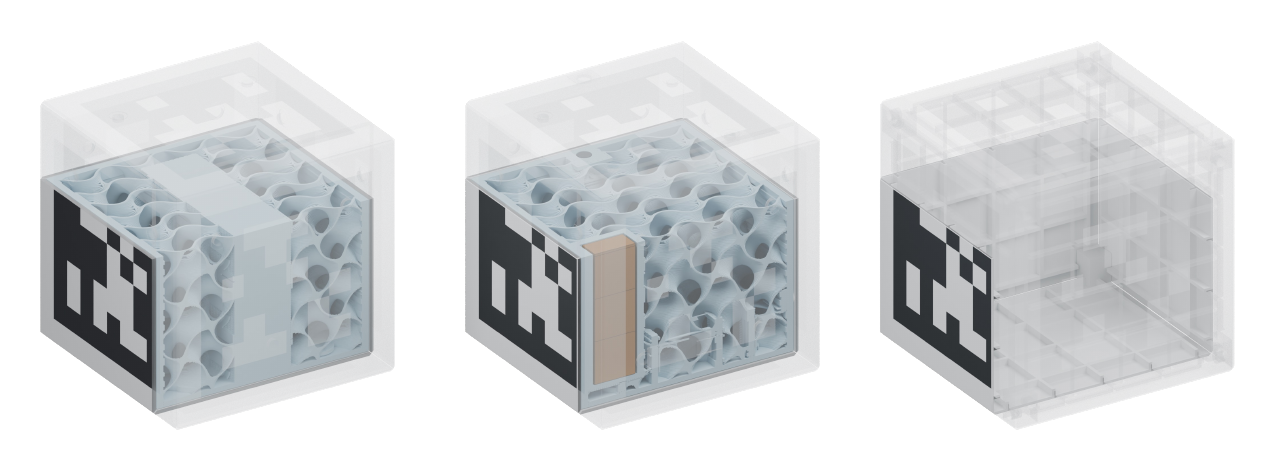}
\caption{\textbf{Real-world test objects.} Three 3D-printed cubes with the same appearance but different physical properties. \textit{Left:} a standard cube with uniform mass. \textit{Middle:} a cube with a metal insert near one corner, which shifts its center of mass. \textit{Right:}  a hollow, lightweight cube, used with low-friction tape on the gripper fingers to reduce friction.}
\label{fig:real_objects}
\end{figure}

\noindent\textbf{Real-World Evaluation.} To test whether our findings carry over to the physical world, we build a real-world setup that mirrors our simulation tasks (Figure~\ref{fig:robot_setup}). It consists of a UR10e arm with a Robotiq 2F-85 gripper and a wrist-mounted F/T sensor. A base camera (RealSense D455) and a wrist camera (RealSense D405) provide visual observations. Note that this arm differs from the UR5e used in simulation. For the target objects, we design three 3D-printed cubes that look identical but have different physical properties (Figure~\ref{fig:real_objects}): a standard cube with uniform mass, a cube with a metal insert near one corner that shifts its center of mass, and a hollow lightweight cube that we pair with low-friction tape on the gripper fingers. Because the cubes look identical, a policy cannot tell them apart from visual cues and must instead infer their physical properties through interaction. In each trial, the robot must throw the cube into a basket placed outside its workspace. We collect demonstrations following the same principles as the \datasetstyle generation pipeline (see Appendix~\textbf{\textcolor{dexorange}{B}}), with 300 demonstrations per object, fine-tune each policy on this data, and evaluate it over 50 rollouts per object. As shown in Table~\ref{tab:real_world}, DynamicVLA fails on every trial, and GR00T N1.7 reaches an average success rate of only 10.7\%. Its performance drops as the physics deviates from the standard cube: 16\% on the uniform cube, 12\% on the shifted-CoM cube, and only 4\% on the lightweight low-friction cube. In contrast, ForceVLA reaches 51.3\% on average. We attribute this gap to its force/torque input, which directly measures the object's weight and inertia once it is grasped. This finding agrees with our ground-truth physics experiment: when a policy has access to physical information, its performance improves substantially. Even so, ForceVLA also degrades on the lightweight low-friction cube (58\% $\rightarrow$ 40\%), where reduced friction causes the object to slip during the throw and a lighter mass makes its trajectory more sensitive to the release velocity. Overall, these results confirm that physical blindness is a key limitation of vision-only policies in the real world, and that physical sensing is a promising direction for addressing it.

\begin{figure}[t]
    \centering
    \includegraphics[width=\columnwidth]{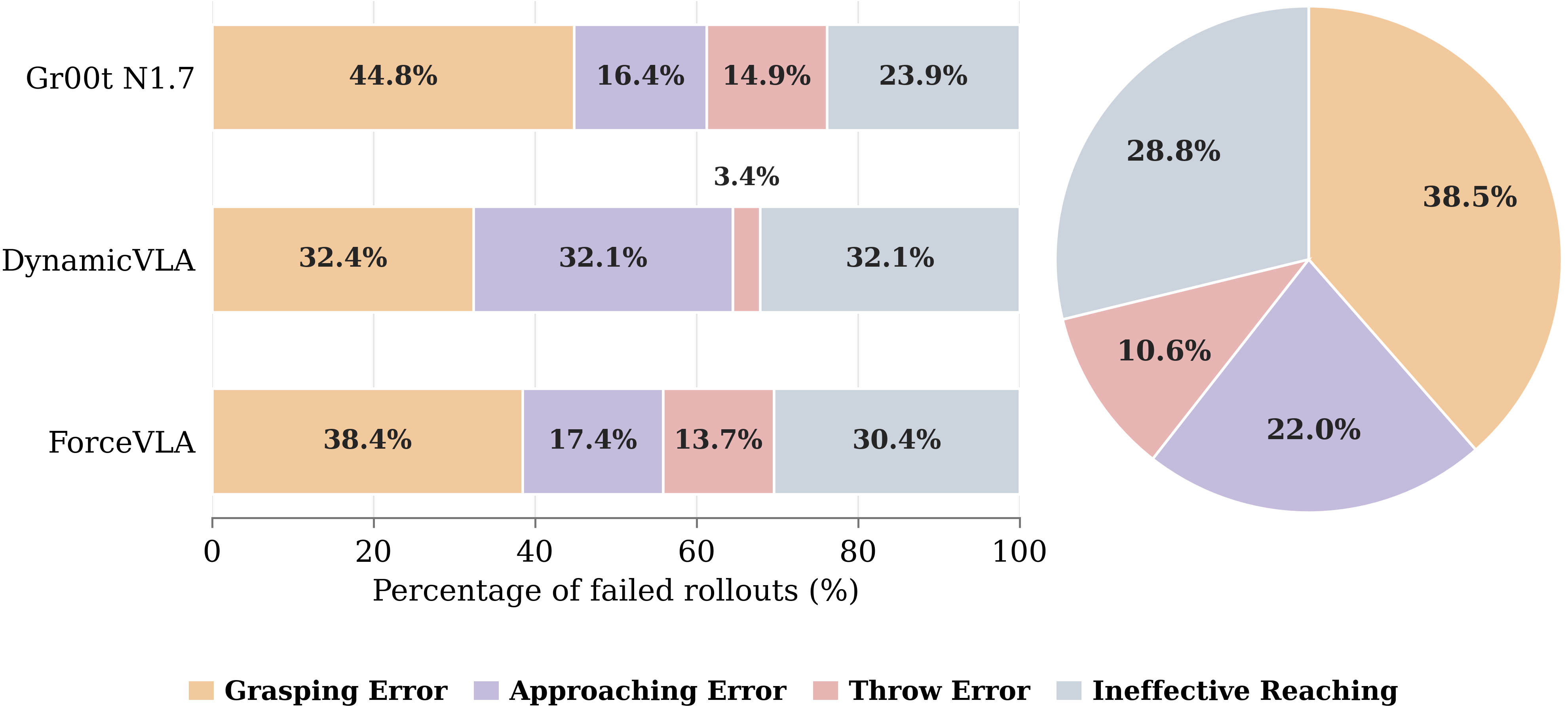}
\caption{\textbf{Failure analysis.} Breakdown of failure modes across all evaluated policies.}
    \vspace{-5mm}
\label{fig:failure_only_measured}
\end{figure}

\noindent\textbf{Failure Analysis.} To understand where current policies break down, we sort every failed rollout into one of four failure modes, ordered by manipulation stage: \textit{ineffective reaching}, where the arm never makes meaningful progress toward the target object; \textit{approaching error}, where the arm reaches the object but arrives in a pose that does not allow a grasp; \textit{grasping error}, where the gripper fails to secure the object or drops it before release; and \textit{throwing error}, where the object is grasped and released but misses the destination. As shown in Figure~\ref{fig:failure_only_measured}, about 89\% of all failures occur before the throw begins, and grasping error is the most common mode (38.5\%), followed by ineffective reaching (28.8\%) and approaching error (22.0\%). These pre-throw failures are consistent with the limited robustness discussed in Section~\ref{sec:overall_results}: under our domain randomization of object poses, camera views, and scene layouts, policies often cannot even complete the grasp. The breakdown also differs across policies. DynamicVLA fails mainly in the earliest stages, with ineffective reaching and approaching errors together accounting for 64.2\% of its failures, which matches its weak SmolLM2 backbone. Its throwing error is only 3.4\% simply because it rarely gets far enough to throw. GR00T N1.7 and ForceVLA reach the object more reliably but still fail mostly at grasping (44.8\% and 38.4\%). Throwing errors make up a smaller share of failures (10.6\% overall), but this reflects how few rollouts reach the throwing stage, not throwing competence: once a throw is attempted, it rarely succeeds. Together, these results suggest that improving physical understanding alone is not enough, and future policies must also become more robust in the reaching and grasping stages that come before it.

\begin{table}[t]
\caption{\textbf{Real-worl experiment.} Top: real-world setup Bottom: success rates by destination type.}
\renewcommand{\arraystretch}{1.25}
\resizebox{\columnwidth}{!}{
\begin{tabular}{lccc}
\toprule
Object & GR00T N1.7~\cite{bjorck2025gr00t} & DynamicVLA~\cite{xie2026dynamicvla} & ForceVLA~\cite{yu2026forcevla} \\
\midrule
Cube 1   & 16\%  & 0\% &  58\%  \\
Cube 2 & 12\%  & 0\% & 56\%  \\
Cube 3   & 4\%  & 0\% & 40\%  \\
\midrule
\textit{Average} & 10.7\%  & 0\% & 51.3\%  \\
\bottomrule
\end{tabular}
}

\label{tab:real_world}
    \vspace{-5mm}

\end{table}

\section{Conclusion \& Discussion}
In this work, we introduced \datasetstyle, a benchmark that evaluates physical reasoning in VLA models through dynamic extrinsic dexterity. Unlike existing benchmarks that focus on gripper--object contact, \datasetstyle requires policies to infer an object's mass, center of mass, and friction and to throw it to a destination beyond the robot's kinematic workspace. Built on a scalable and fully automatic pipeline, it provides 35K trajectories across nine tasks and 100 scenes, covering uniform, irregular, and dynamic-mass objects with static, multiple, and moving destinations. Our evaluation of three representative VLAs, GR00T N1.7, DynamicVLA, and ForceVLA, shows that current policies are far from reliable on this problem, with overall success rates below 3\%. We identify two main bottlenecks: limited robustness of pretrained models, which causes most failures to occur before the throw begins, and physical blindness, since policies cannot infer properties that are not directly observable. Giving policies ground-truth physical properties roughly doubles their success rate, and experiments on a humanoid robot show the same failure patterns, which suggests these limitations are not tied to a single embodiment. We hope \datasetstyle will serve as a testbed for physically grounded robot learning and motivate new designs, such as explicit physical property estimation, VLA training with reinforcement learning, and richer multi-sensory inputs.

\noindent\textbf{Limitations.} Our study has several limitations. First, \datasetstyle focuses on a single dynamic skill, throwing, and relies on simplified physical assumptions, randomizing only mass, center of mass, and friction. Even so, current policies already fail at this level of difficulty, which shows a clear lack of physical intelligence. Extending the benchmark to more skills, such as pushing, sliding, and pouring, and to richer physics, such as elasticity, deformable objects, and aerodynamic effects, would make it more comprehensive. Second, we evaluate only three representative VLA policies. Including a wider range of models would make the conclusions stronger. Third, we discuss alternative approaches only briefly. Methods such as in-context learning, which can adapt from a few interactions, and reinforcement learning, which can discover physical properties through trial and error, may be better at probing physical properties, and we leave a systematic study of them to future work.

\noindent\textbf{Broader Impact.} \xspace \datasetstyle aims to advance robots that understand and act on the physical properties of the objects they handle. This ability is important for safe, reliable deployment in homes, warehouses, and factories, where objects vary widely in mass, shape, and friction. By exposing the physical blindness of current VLA policies, our benchmark can help the community build models that fail less often and less unpredictably in real-world settings. Because \datasetstyle is fully synthetic and generated automatically, it also lowers the cost of studying dynamic manipulation, which is otherwise expensive and risky to collect on real hardware. At the same time, throwing is a high-speed, high-energy behavior, and a policy that misjudges an object's properties could release it on an unintended trajectory and harm people or damage property nearby. Policies trained or validated on \datasetstyle should therefore not be deployed without additional safety measures, such as restricted workspaces, speed and force limits, and human oversight. Moreover, strong performance in simulation does not guarantee safe behavior in the real world, so results on \datasetstyle should be complemented by careful physical testing before deployment.

\section*{LLM Usage Statement}
Claude Opus was used to assist with language editing. The authors are responsible for verifying all mathematical derivations, experimental results, references, and scientific claims in the final manuscript.

{
    \small
    \bibliographystyle{ieeenat_fullname}
    \bibliography{main}
}





\end{document}